\documentclass[letterpaper, 10 pt, conference]{ieeeconf}  

\IEEEoverridecommandlockouts                              

\usepackage{bmpsize}\usepackage[utf8]{inputenc}
\usepackage{graphicx}\usepackage[utf8]{inputenc}
\usepackage{amssymb,amsmath}
\usepackage{bm}
\usepackage[english]{babel}

\makeatletter
\renewcommand*\env@matrix[1][\arraystretch]{%
	\edef\arraystretch{#1}%
	\hskip -\arraycolsep
	\let\@ifnextchar\new@ifnextchar
	\array{*\c@MaxMatrixCols c}}
\makeatother

\usepackage{float}
\usepackage{amssymb}
\usepackage{soul}
\graphicspath{{Images/}}
\usepackage[colorlinks={true},linkcolor={black},citecolor={black}, urlcolor={black}]{hyperref}
\emergencystretch\maxdimen
\title{\LARGE \bf
Adaptive Tracking Control of Soft Robots using Integrated Sensing Skin and Recurrent Neural Networks
}

\author{Lasitha Weerakoon$^{1}$, Zepeng Ye$^{1}$, Rahul Subramonian Bama$^1$, Elisabeth Smela$^{1}$, Miao Yu$^{1}$ and Nikhil Chopra$^{1}$
\thanks{$^{1}$Department of Mechanical Engineering, University of Maryland, College Park, MD 20742, USA.
        {\tt\small lasitha@umd.edu, hudsonmars01@gmail.com, sb.rahul@hotmail.com, smela@umd.edu, mmyu@umd.edu, nchopra@umd.edu}}%
}

\begin{document}

\maketitle
\thispagestyle{empty}
\pagestyle{empty}
\begin{abstract}
In this paper, we study integrated estimation and control of soft robots. A significant challenge in deploying closed loop controllers is reliable proprioception via integrated sensing in soft robots. Despite the considerable advances accomplished in fabrication, modelling, and model-based control of soft robots, integrated sensing and estimation is still in its infancy. To that end, this paper introduces a new method of estimating the degree of curvature of a soft robot using a stretchable sensing skin. The skin is a spray-coated piezoresistive sensing layer on a latex membrane. The mapping from the strain signal to the degree of curvature is estimated by using a recurrent neural network. We investigate uni-directional bending as well as bi-directional bending of a single-segment soft robot. Moreover, an adaptive controller is developed to track the degree of curvature of the soft robot in the presence of dynamic uncertainties. Subsequently, using the integrated soft sensing skin, we experimentally demonstrate successful curvature tracking control of the soft robot.
\end{abstract}

\section{Introduction} \label{sec:intro}

Soft robots are defined as systems fabricated from materials that have low elastic moduli, in the range of those of biological materials ($10^4$–$10^9$Pa) \cite{rus2015design} or of elastomers. The compliance and dexterity of soft robots can be utilized for effective manipulation in unstructured environments. The robust and agile environmental manipulation by animals, such as the octopus's varied use of tentacles and elephant's dexterous trunk \cite{majidi2014soft, SurveyPaper_george2018control}, have also inspired the development of soft robots. Several applications ranging from underwater and space operations to minimally invasive surgeries have been identified for soft robots \cite{SurveyPaper_george2018control,webster}.


An important recent focus of the soft robotics community has been the development of integrated sensors for soft robotic perception (e.g., \cite{2019_Thuruthel_soft, 2020_Truby_distributed}). Integrated sensing would potentially enable a soft robot to perceive the world without external sensors. The sensory signals acquired from integrated sensors can then be utilized for state estimation and in closed loop control. Several  methods have been proposed for developing integrated sensors for soft robots \cite{2019_Thuruthel_soft, truby2018soft}.  However, only a few sensing technologies have been demonstrated that are readily amenable to closed loop dynamic control for a wide range of soft robots \cite{2020_Truby_distributed}.


\begin{figure}[t]
	\centering
	    \includegraphics[width=0.47\textwidth]{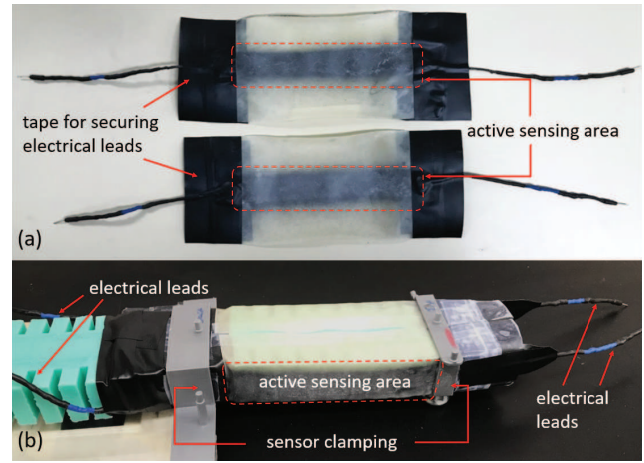} \vspace{-3pt} \\
	\caption{The soft sensing skins (a) and the soft robot retrofitted with the sensors (b)} \vspace{-12pt}
	\label{Fig:new_skin_and_robot_figure}
\end{figure}

In this paper, we investigate adaptive tracking control of a planar soft robot with a simple sensing skin that can be easily retrofitted to estimate the degree of curvature. While this study considered a planar single-segment soft robot capable of bi-directional bending with a constant curvature along the length of the segment, the proposed advances could also be utilized for multi-segment 3D soft robots. The sensing skin consisted of a piezoresistive sensing layer spray coated onto a latex membrane \cite{2018_Ying_Chen_compliant, barnett2017targeted}. A strip-shaped sensing area was created, and electrical leads were attached at either end. The sensing skin and the soft robot retrofitted with the sensors are shown in Fig.\ref{Fig:new_skin_and_robot_figure}. A data driven model, namely a  long short term memory (LSTM) network, a special recurrent neural network (RNN), was used to determine the relationship between the sensor signals and the degree of curvature. Both uni-directional bending and bi-directional bending were investigated. We successfully demonstrate the utilization of the proposed integrated sensing strategy in an adaptive control framework \cite{adaptive_slotine} for dynamic tracking control of soft robots in experiments. The adaptive controller was developed to track the degree of curvature of the soft robot assuming uncertainty in the dynamic parameters of the soft robot.

To the best of our knowledge, we demonstrate the first steps toward utilizing retrofitted soft sensor skins for degree of curvature estimation in adaptive tracking control of soft robots for bi-directional bending.



The rest of the paper is organized as follows. Related work is discussed in Section \ref{sec:related_work}. The dynamic model of the soft robot and the adaptive tracking control framework are discussed in Section \ref{sec:Theory}. Section \ref{sec:methods} introduces the soft sensing skin and the soft robot, and discusses degree of curvature estimation using integrated sensing skins. The experimental results for curvature tracking control using the integrated sensing are presented in Section \ref{sec:experiments} and the results are discussed in Section \ref{sec:discussion}. Finally, Section \ref{sec:conclusion} summarizes the paper and outlines potential future directions.

\section{Related work} \label{sec:related_work}
Over the past decade, a considerable amount of work on developing embedded sensing for soft robots has emerged. Some studies demonstrated the use of commercially available flex sensors embedded in a soft robot to measure the bending of the body. In one study \cite{gerboni2017feedback},  the integration of commercial flex sensors within a soft bending module actuated by pressure-driven fluidic actuators was attempted. Commercial flex sensors were used in another study \cite{2018_Elgeneidy_bending} to estimate bending angle through a data-driven approach, in which the  bending angle control was achieved by utilizing the predicted angle in a classical PID controller (heuristically tuned, rather than model-based). Flex sensors were also used in soft elastomer composite actuators for bending angle estimation \cite{tang2019novel, 2019_Tang_Human_wearable} . The estimates were then utilized in learning based control frameworks. In another work \cite{wang2019parameter}, a flex sensor was used for bending angle estimation, which was then employed in a back-stepping control algorithm for bending angle tracking. More recently,  a data driven model was proposed to estimate the bending angle from a commercially available flex sensor as well as the pressure data \cite{mohamed2020proposed}. The paper demonstrated the use of the estimates in a model-free static controller for bending angle control. 

One drawback of the commercial flex sensors is that they stiffen the soft bodies since the flex sensors are not as soft as the soft robot body \cite{2019_Thuruthel_soft, yang2017innovative}. Specifically, the flex sensors bend but they do not stretch.  Therefore, they are embedded in the non-stretching region at the center of the robot segment \cite{yuen2018strain}. Flex sensors could not be used in extensible soft robots because of their lack of stretchability. Thus, some groups have focused on developing soft embedded sensors that do not impact the mechanical compliance of the soft robots. Such embedded sensors for estimating soft robot position, actuation pressure, and force sensing have been fabricated recently \cite{yang2017innovative, yuen2018strain, tiziani2017sensorized}. A method to fabricate soft somatosensitive actuators by embedding 3D printed ionically conductive gels was proposed in \cite{truby2018soft}. In \cite{kim2020sustainable} a McKibben-type actuator with an embedded soft sensor was fabricated using a self-coagulating conductive Pickering emulsion and was used in closed loop control for slow movements with considerable error. In \cite{2020_Zhou_differential_CCPF}, a differential sensing method for the application of soft robot angle sensing using an embedded coiled conductive polymer fiber was proposed. A closed-loop multidimensional angle control system based on PID control using the differential sensing method was then developed to verify the sensing performance. The review paper \cite{zolfagharian20203d} discusses soft pneumatic actuators fabricated entirely with additive manufacturing methods and suggests learning based control for soft robots with self-sensing capability.

Recent efforts in embedded sensing technology have used polydimethylsiloxane (PDMS) filled with carbon nanotubes (cPDMS) \cite{2019_Thuruthel_soft,shih2017custom}. The resistance of these polymers increases with strain \cite{liu2009patterning}. By embedding cPDMSs in the soft robot body and measuring the resistance of these areas, the bending of the soft body could be estimated. In \cite{2019_Thuruthel_soft}, the authors discussed a strategy for data-driven multi-modal sensing, namely the robot tip position and exerted force at the tip, using a cPDMS embedded sensor. The fabrication of cPDMS soft skins and their use for tactile sensing for haptic visualization was discussed in \cite{shih2017custom} .

The authors in \cite{2020_Truby_distributed} used off-the-shelf conductive silicone elastomer sheets laser cut into Kirigami patters and bonded to the soft robot skin as soft piezoresistive silicone sensors. Using these sensors, the steady state 3D configuration of the soft robot was predicted using a trained RNN. This strategy has been used for developing data–driven disturbance observers for estimating external forces on soft robots \cite{2020_Della_Santina_data_driven}.

\section{Soft robot control framework}\label{sec:Theory}

In this section we introduce the nonlinear dynamic model of the soft robot assuming the piecewise constant curvature (PCC) hypothesis \cite{della2019model}. Subsequently, the adaptive control framework for curvature tracking will be developed.

\subsection{Soft robot model}
While the sensing skin and the experimental results are developed for a single segment soft robot, the approach is scalable, and hence the general multi segment dynamics and control strategy is discussed here. The dynamics of the soft robot are formulated as a Lagrangian system through a dynamically consistent \textit{Augmented Formulation} using the methods introduced in \cite{della2019model}. We consider a PCC soft robot with $n$ inextensible constant curvature (CC) segments with masses $m_i$ and lengths $L_i$. Using the \textit{Augmented Formulation}, the soft robot is modeled as a sequence of revolute and prismatic (R and P) joints, matching the kinematic and dynamic properties of the soft robot to an approximated rigid robot system. 

Following \cite{della2019model}, in the absence of external wrenches the complete dynamics of the soft robot are represented as an approximated rigid robot evolving on the degree of curvature space $q_s(t) \in R^n$:
	{\begin{align}
		M_s(q_s) \ddot{q_s} + C_s(q_s,\dot{q_s})\dot{q_s}+D_s\dot{q_s} +K_s q_s +G_s(q_s) = \tau_s,
		\label{soft_model}
		\end{align}
}where {\small$M_s(q_s)\in R^{n \times n}$} is the equivalent inertial matrix, {\small$C_s(q_s,\dot{q}_s)\dot{q}_{s} \in R^{n} $} represents the equivalent centrifugal and Coriolis terms, {\small$G_s(q_s) \in R^n$} is the equivalent gravitational torque vector, and {\small$\tau_s \in R^{n}$} is the generalized torque vector. Here, damping and stiffness matrices, {\small$D_s, K_s \in R^{n \times n}$} respectively, are introduced to incorporate the compliance of the soft robot. The interested reader is referred to \cite{della2019model} for the complete derivation.

\subsection{Adaptive controller design}
The motivation for developing an adaptive controller is that the estimated model for the soft robot in the form of a PCC segment based rigid robot manipulator is not exact. Also the parameters of the model are not known precisely. Therefore a control mechanism that adapts the parameters as the soft robot operates would be beneficial for good performance. It should be noted that the main objective of the controller is to track the desired curvature. If the PCC model (\ref{soft_model}) is viewed as a rigid manipulator model, we can apply classical methods developed for rigid robots as shown in \cite{adaptive_slotine} to develop the adaptive tracking controller.

Define the degree of curvature error vector as $\tilde{q}_s(t) = q_s(t) - q_d(t)$ where $q_d(t)$ is the desired curvature. Define the virtual reference trajectory $\dot{q}_r(t) = \dot{q}_d(t) - \lambda \tilde{q}_s(t)$ and let $s(t)=\dot{\tilde{q}}_s(t) - \lambda \tilde{q}_s(t)$, where $\lambda$ is a positive definite parameter matrix which needs to be tuned. 

Denote the \textit{equivalent parameter vector} of the model as $\Theta_s$, whose elements are combinations of the variables $m_i$, $L_i$, $K_s$, and $D_s$. Note that the $K_s$ and $D_s$ terms will be explicitly included in $\Theta_s$. Using the properties of Lagrangian systems, define the regressor {\small $\left(Y_s\left(q_s,\dot{q}_s,\dot{q}_r, \ddot{q}_r \right)\right)$} and parameter { $\left(\Theta_s \right)$} vector pair for the augmented soft robot model \cite{spong2006robot}, \vspace{-5pt}
\begin{align*}
	Y_s {\Theta}_s = {M}_s \ddot{q}_r + ({C}_s  +  {D_s})\dot{{q}}_r + {K_s} q_s + {G}_s.
	\vspace{-8pt}
\end{align*}
The estimated equivalent parameter vector is denoted by $\hat{\Theta}_s$, and hence the estimation error is defined as $\tilde{\Theta}_s = \hat{\Theta}_s - \Theta_s$.
Now we propose the control law \vspace{-5pt}
\begin{align}
\tau = Y_s \hat{\Theta}_s - K_D s,
\vspace{-8pt}
\label{eq:adaptiveController}
\end{align}
where $K_D$ is a gain term that needs to be tuned. The adaptation law with the positive definite adaptation gain matrix $\Gamma$ is \vspace{-5pt}
\begin{align}
\dot{\hat{\Theta}}_s = - \Gamma Y_s^T s.
\vspace{-5pt}
\label{eq:adaptationgain}
\end{align}
\vspace{-15pt}\\
The stability of the designed controller (\ref{eq:adaptiveController})-(\ref{eq:adaptationgain}) can be demonstrated using Lyapunov analysis~ \cite{adaptive_slotine, spong2006robot}.

\section{Integrated sensing} \label{sec:methods}

In this section we discuss the proposed method for degree of curvature estimation using soft sensing skins retrofitted onto a soft robot. First, the soft sensing skin \cite{2018_Ying_Chen_compliant, barnett2017targeted} is described and the soft robot is characterized. Then the experimental setup is introduced and the approach for degree of curvature estimation is then discussed.

\subsection{Soft sensing skin}

The soft sensors used in this work were fabricated by spray coating a stretchable piezoresistive sensing layer in the shape of a strip onto a latex membrane. The coating consisted of a latex host filled with exfoliated graphite (EG).  Electromechanical connections to the sensing layer were made using carbon fiber yarn attached with the same latex/EG solution serving as a ``glue" \cite{2018_Ying_Chen_compliant}.  A top coating of latex was added to protect the sensing layer from mechanical damage. Further, adhesive tape was used at the ends of the sensor strips to protect the electromechanical connections of the sensor. The carbon fiber yarn was joined to a copper wire by winding the copper wire around the carbon fiber yarn and then using a heat-shrink tubing to secure the joint.

Upon stretching the sensor, the resistance of the film increases. The response of these sensors is substantially linear over a large range of strain \cite{chen2016stretchable}, and the gauge factor (GF), or sensitivity, is on the order of 10 ($\Delta$R/R = GF*$\epsilon$, where $\Delta$R is the change in resistance, R is the original resistance, and $\epsilon$ is the strain).  

Two stretchable sensing skins were retrofitted over the soft robot, one on each side, to measure strain when the segment was bending. Fig. \ref{Fig:new_skin_and_robot_figure} (a) shows an image of the soft sensing skins. Fig. \ref{Fig:new_skin_and_robot_figure} (b) shows the skins over the outer surface of an actuated segment, mechanically held in place using rectangular rings around the un-actuated ends of the segment; these rings are typically used to hold the markers for the motion capture system, and here served both functions.  Two voltage divider circuits were used to measure the resistance of the sensing skins, and an Arduino board serially transmitted these as analog signals, which are referred to as ``raw strain signals''.


\begin{figure}[t]
	\centering
	\begin{tabular}{c c}
        \includegraphics[height=0.2\textwidth]{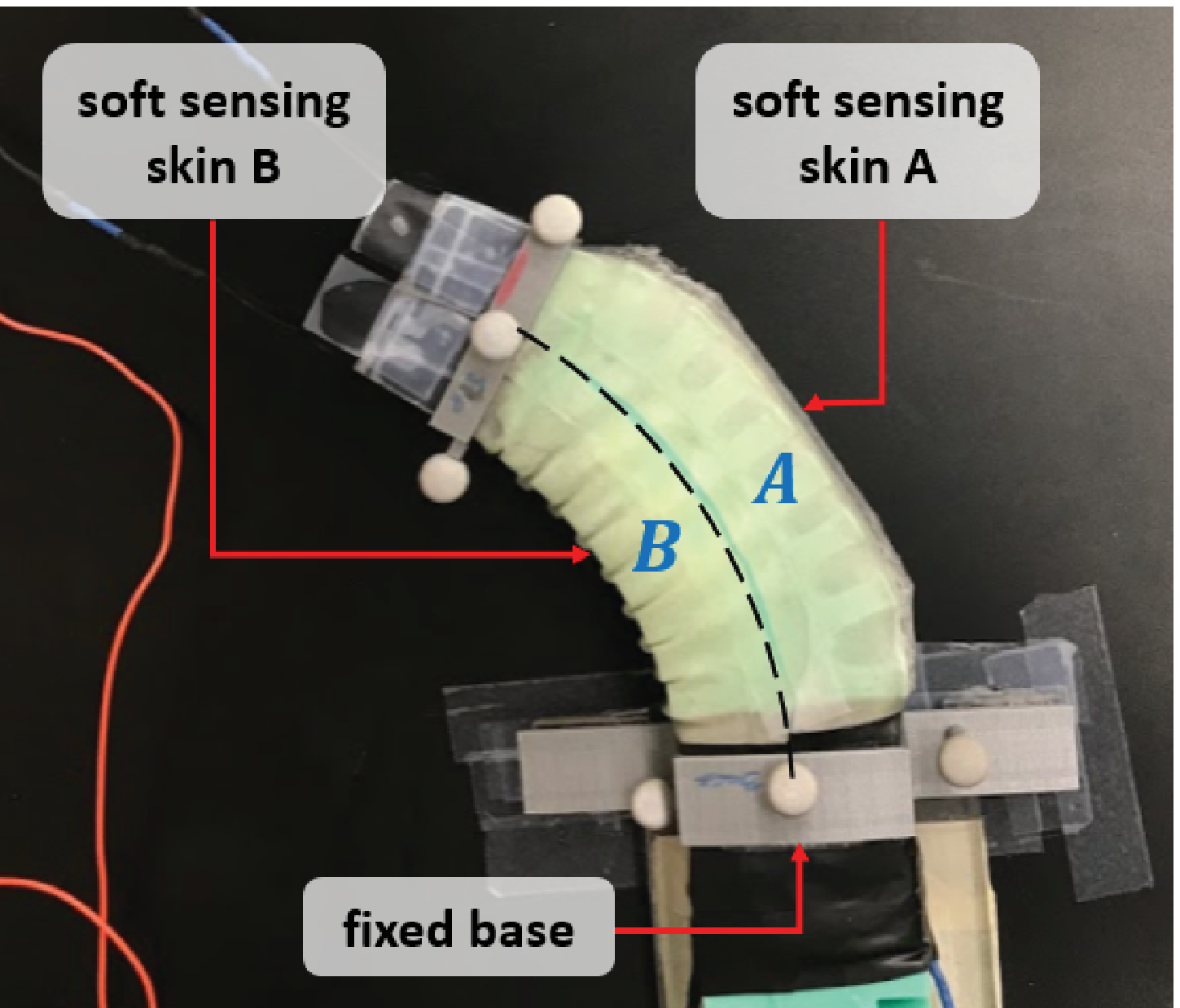} & 
        \includegraphics[height=0.2\textwidth]{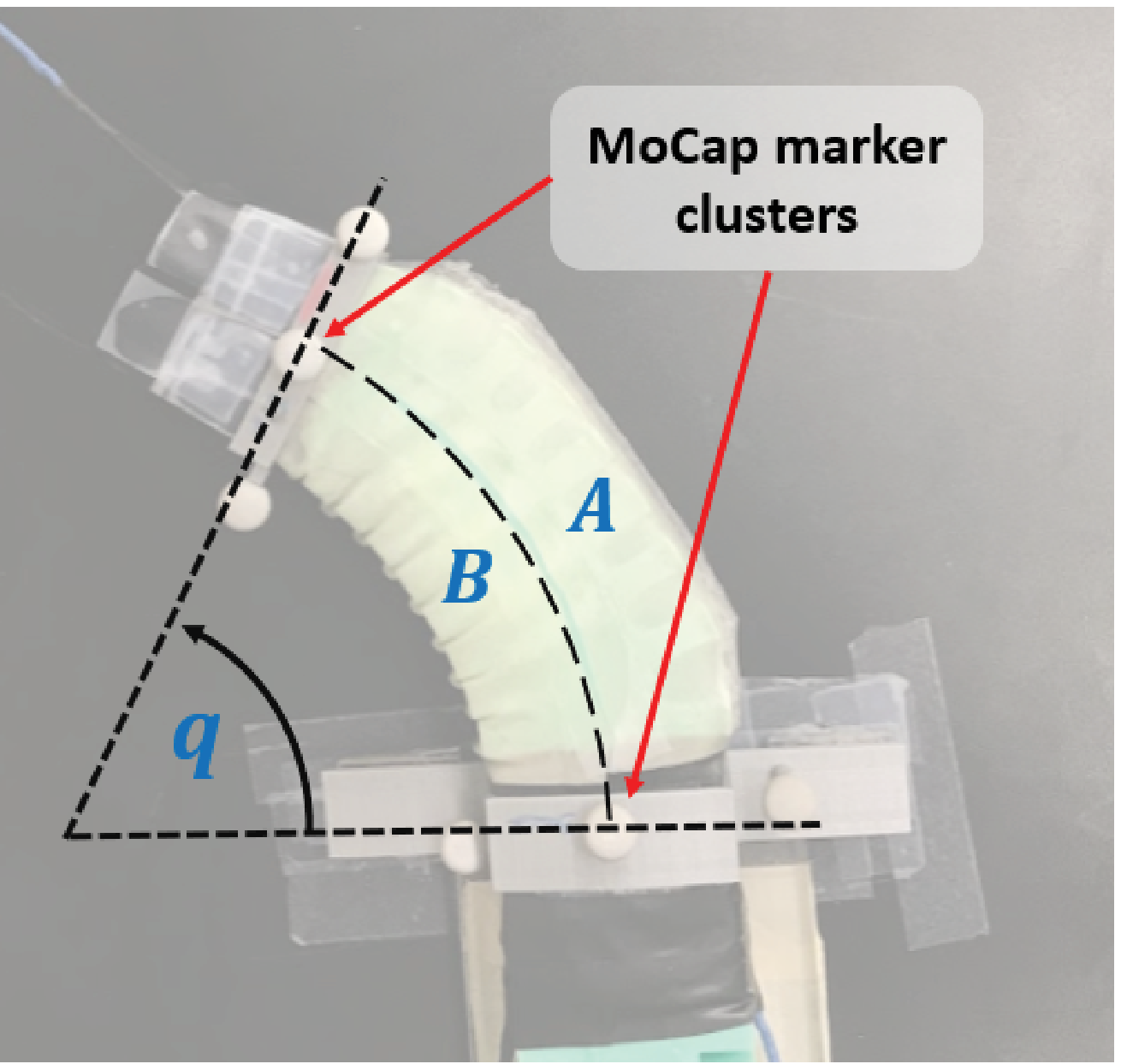} \\
        \footnotesize{(a)} & \footnotesize{(b)} \vspace{-3pt}
		
	\end{tabular}
	\caption{Soft robot retrofitted with the stretchable sensing skins. The \textit{compartments} of the soft robot and the sensing skins are labeled in panel (a). Panel (b) shows the degree of curvature and the markers placed for the motion capture (MoCap) system.} \vspace{-15pt}
	\label{Fig:soft skin} 
\end{figure}

\subsection{The soft robot}
The soft robot used in this work is a planar bi-directional pleated type soft segment which was fabricated following the methods outlined in \cite{Marchese2015_recipe}. The same soft robot in~\cite{2020_Lasitha_teleop_ACC} by the authors was utilized, except that only the distal segment was actuated. The base of this distal segment was fixed and the segment was constrained to move on a horizontal table. We refer to this actuated single-segment as the soft robot for this study. Ball transfers were used underneath, near the robot tip, to reduce friction when bending. The segment had two {\it compartments}, named $A$ and $B$ as shown in Fig. \ref{Fig:soft skin}, that were individually actuated pneumatically, and were assumed to deform with a constant curvature along the length of the segment under the applied pressure. The {\it middle layer} of the segment {(the joint between the two chambers)} was inextensible due to the restrained material layer. 

The segment length along the inextensible \textit{middle layer} was measured to be $L_{1} = 124$ mm. The segment mass $m_1$ is uncertain due to the retrofitting the soft sensing skin on the original design, for which the segment mass $0.110$ kg was measured prior to joining the segments together. 

The soft robot was actuated using a pneumatic controller unit based on an open source hardware platform \cite{softroboticstoolkit}, where the actuation signals were serially transmitted to the control board (Arduino Mega). An external compressor supplied compressed air to the actuation unit at a constant pressure of 20 psi. The air pressure in the soft segment was regulated by a pulse-width modulation (PWM) at 100 Hz. The control input calculated by the controller, in terms of a {\it torque}, was converted to a PWM signal for the segment using an appropriate mapping. At a given time instance, depending on the sign of the torque, only one compartment out of the two in the segment was actuated. A positive torque commanded compartment $A$ of the segment, while a negative torque commanded compartment $B$.

\begin{figure}[t]
	\centering
		\includegraphics[width=0.4\textwidth]{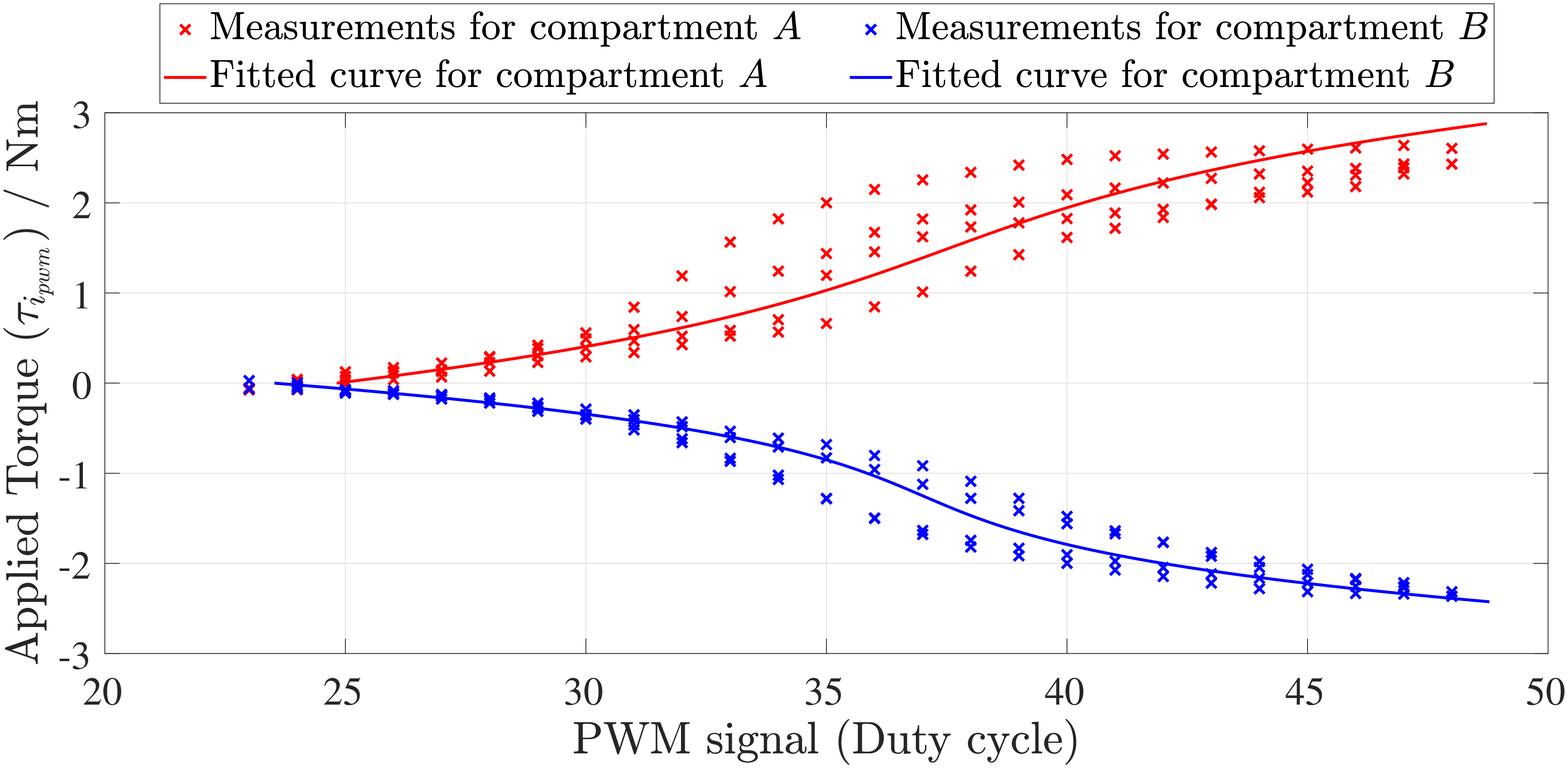}
		\vspace{-8pt}\\
	\caption{Relationship between PWM signal and applied torque}
	\label{Fig:curvefit1}  \vspace{-12pt}
\end{figure}

The torque-to-PWM signal mapping was identified by a curve fitting process for each of the two compartments. Considering the dynamic model of the soft robot (\ref{soft_model}), a step input of $\tau_{s} = \tau_{\text{pwm}}$ resulted in a steady state represented by $\tau_{{\text{pwm}}} = K_s \: \theta_{{\text{pwm}}}$. Here $\theta_{{\text{pwm}}}$ is the steady state degree of curvature of the segment and $K_s$ is the torsional stiffness. Sending commands to only one compartment of the segment, the steady state degree of curvatures ($\theta_{{\text{pwm}}}$) was recorded for different PWM signal values. Then the applied equivalent torque, $\tau_{{\text{pwm}}}$, was calculated hypothesizing the torsional stiffness of the segment to have a nominal value of $K_s = 1$ Nm/rad. Finally, a third order polynomial curve fit was performed using this data for each compartment of the segment to obtain the mapping from torque to PWM signal. Fig.\ref{Fig:curvefit1} illustrates the relationship between the PWM signals and the equivalent applied torque for the individual compartments of the soft robot.

The nominal value of the torsional damping $D_s=0.2$ Nms/rad was then calculated using a system identification process by applying a unit torque.

\subsection{The experimental setup}

\begin{figure}[t]
	\centering
		\includegraphics[width=0.425\textwidth]{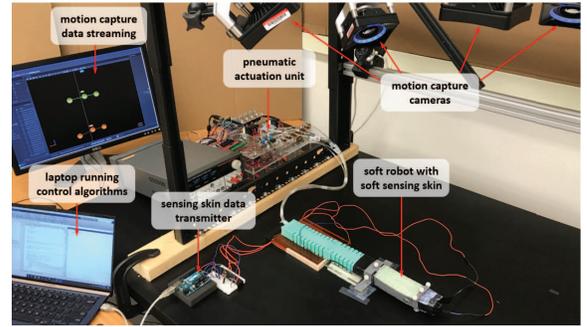}
		\vspace{-8pt}\\
	\caption{The experimental setup}  \vspace{-15pt}
	\label{Fig:setup}
\end{figure}

The experimental setup, shown in Fig. \ref{Fig:setup}, consisted of the single-segment soft robot retrofitted with the sensing skin, the pneumatic actuation unit, an Arduino board to acquire the sensor skin strain signal, an OptiTrack motion capture system for ground truth measurements, and an i7 16GM RAM Windows 10 laptop to train the neural networks and run the control algorithm on MATLAB 2019a. Before each experiment, either data collection or control, two cycles of inflating and deflating each compartment for 5 s intervals were carried out to eliminate \textit{first cycle effects} in the elastomers.

\subsection{Degree of curvature estimation}

We discuss the degree of curvature estimation for two cases. First, uni-directional bending of the soft robot with the utilization of a single soft sensing skin on one side. Second, bi-directional bending utilizing both skins on two sides. We assume that the soft robot segment has the same curvature along its length. A data driven approach was used to identify the relationship between the strain signals from the sensors and the degree of curvature. Specifically, we utilized a Recurrent Neural Network (RNN) named Long Short Term Memory (LSTM) network \cite{LSTM_hochreiter1997long} to learn the time series mapping. 

For both the scenarios the same network architecture designed using the MATLAB Deep Learning Toolbox was used. We used a dropout layer with a rate of 0.1 after the input layer to prevent over-fitting and make predictions more robust to noise. Next, an LSTM layer was used. The number of hidden units for the this layer was selected to be as small as possible to prevent overfitting via a validation set. Then a fully connected layer was added to compute the outputs. 


For the uni-directional bending case, the actuator signal (PWM signal) for the actuated \textit{compartment A} and the raw strain signal from the soft sensing skin A were the inputs to the network. For the bi-directional bending case, the actuator signals for both compartments and the raw strain signals from the two sensing skins were the inputs. The networks' output  was  the  degree  of curvature. The degree of curvature measured using the motion capture (MoCap) system was used as the ground truth when training the networks. Both the networks were trained using the Adam optimizer. L2 regularization with the default value $(0.0001)$ was used. Further, two separate validation sets were used: an $\alpha$-validation set with a frequency of 25 and patience of 5 to minimize over-fitting by early stopping, and a $\beta$-validation set  for selecting the number of hidden units for the LSTM layer by manually inspecting the root mean squared error (RMSE) value for the $\beta$-validation set after the training had stopped.

\subsubsection{Uni-directional bending}

\begin{figure}[t]
	\centering
	\begin{tabular}{c}
		\includegraphics[width=0.48\textwidth]{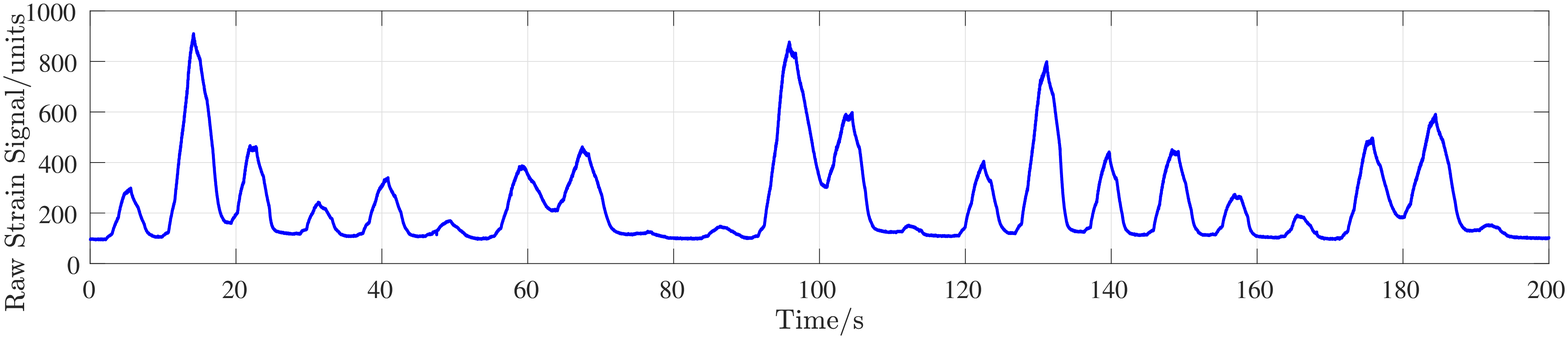}\vspace{-5pt}\\
		\vspace{5pt}
	    \footnotesize{(a) Raw strain signals}\\
	    \vspace{-5pt}
		\includegraphics[width=0.48\textwidth]{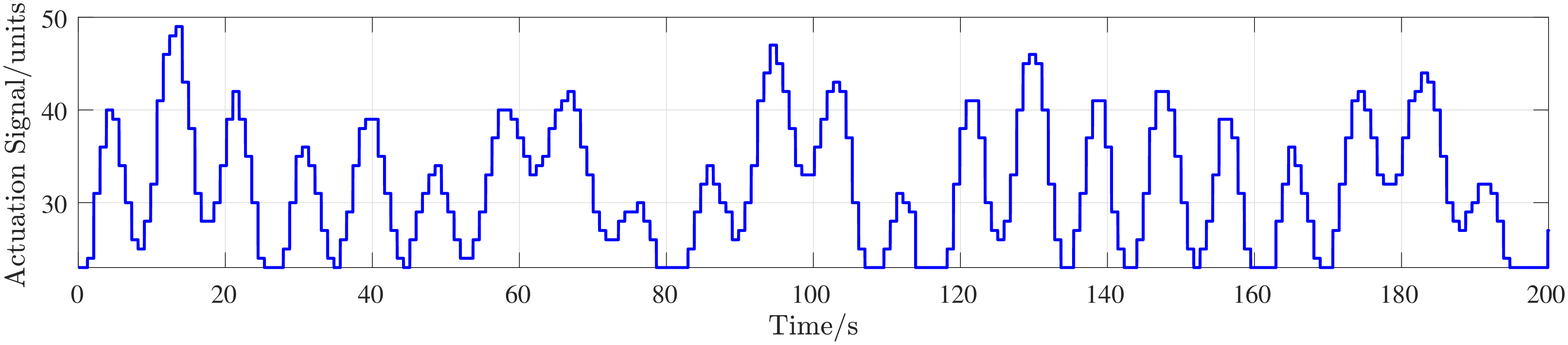}\\
		\vspace{5pt}
		\footnotesize{(b) Actuation signals}\\
		\vspace{-5pt}
		\includegraphics[width=0.48\textwidth]{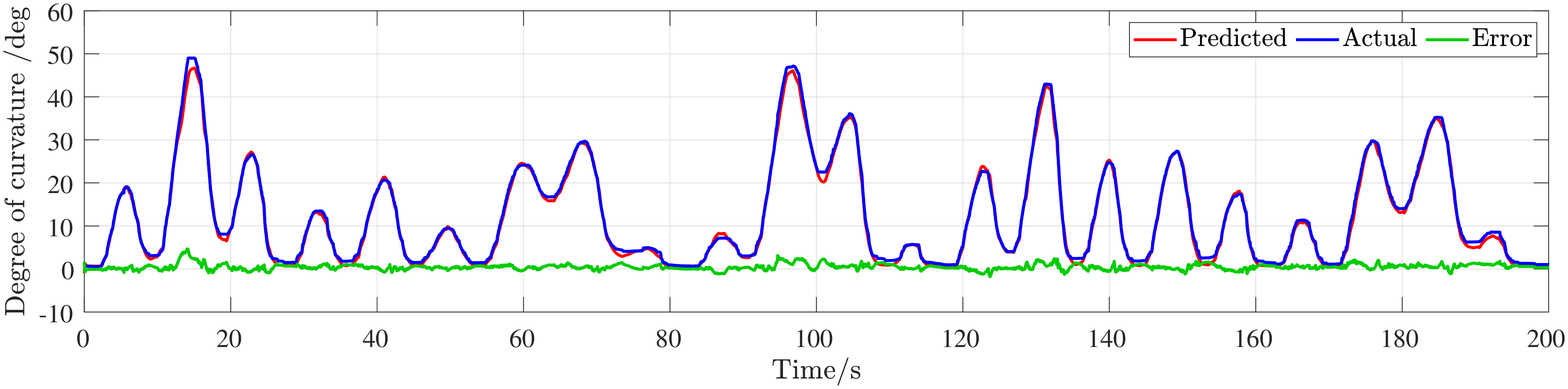}\\
		\vspace{5pt}
		\footnotesize{(c) Test set prediction from learned model} \vspace{-5pt}
	\end{tabular}
	\caption{Learned model test set performance: Uni-directional bending }
	\label{Fig:ML_model_single} \vspace{-12pt}
\end{figure} 

For testing the uni- directional bending, only the \textit{compartment A} of the soft robot was actuated, and the strain signals from the soft sensing skin retrofitted on the \textit{compartment A} were used. Eight experiments were conducted to collect data for training. A random actuation pattern was generated at a rate of approximately $1$ Hz to actuate the soft robot and each experiment was run for a period of $2-3$ minutes. For each experiment, the actuation signal, the strain signal, and the actual degree of curvature were recorded at a rate of $85$ Hz. The collected data were joined together later to constitute the total data set, which consisted of $115,410$ data points. This data set was then divided into a training set of $80,786$ points and two validation sets, $\alpha , \beta$-validation sets, of $17,312$ points each, from which the network was trained. The optimum number of hidden units for the LSTM layer was 30.

Once the RNN was learned, the degree of curvature estimation performance was evaluated in real time for a random actuation pattern. The predicted degree of curvature by the LSTM network superimposed with the actual values for this experiment are illustrated in Fig.\ref{Fig:ML_model_single} along with the raw strain signals and the actuation signals. The RMSE for this test experiment was $0.86^{\circ}$.


\subsubsection{Bi-directional bending}

\begin{figure}[t]
	\centering
	\begin{tabular}{c}
		\vspace{-5pt}
		\includegraphics[width=0.48\textwidth]{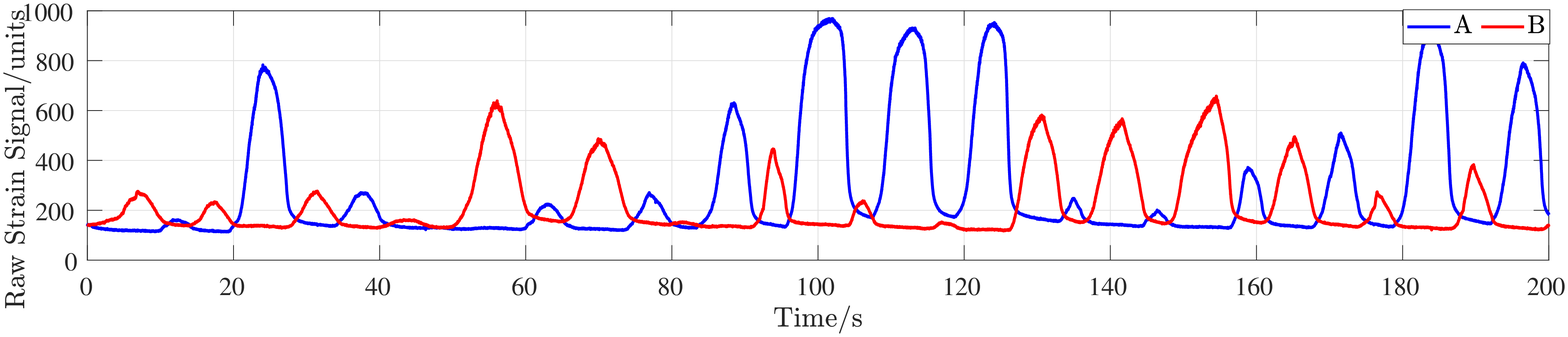}\\
		\vspace{5pt}
	    \footnotesize{(a) Raw strain signals}\\
	    \vspace{-5pt}
		\includegraphics[width=0.48\textwidth]{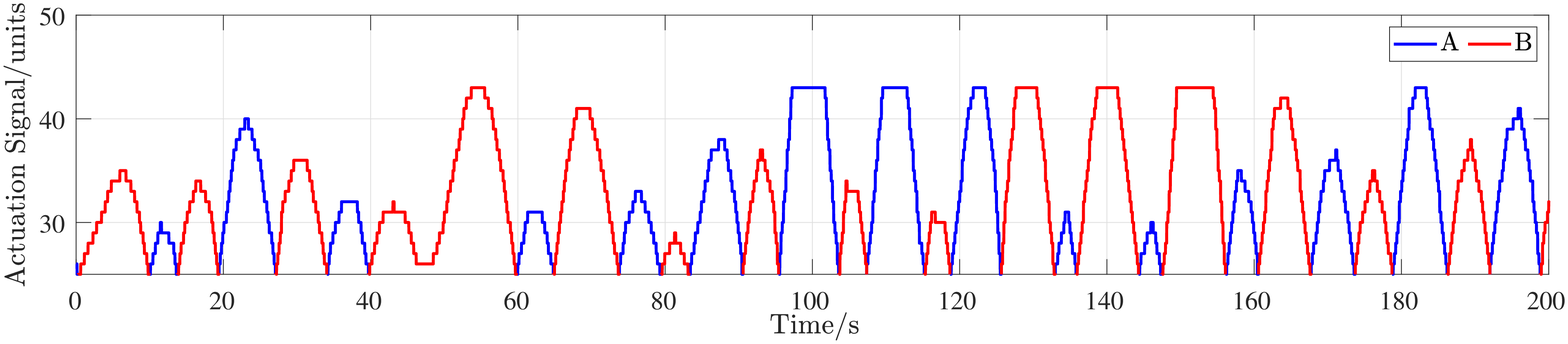}\\
		\vspace{5pt}
		\footnotesize{(b) Actuation signals}\\
		\vspace{-5pt}
		\includegraphics[width=0.48\textwidth]{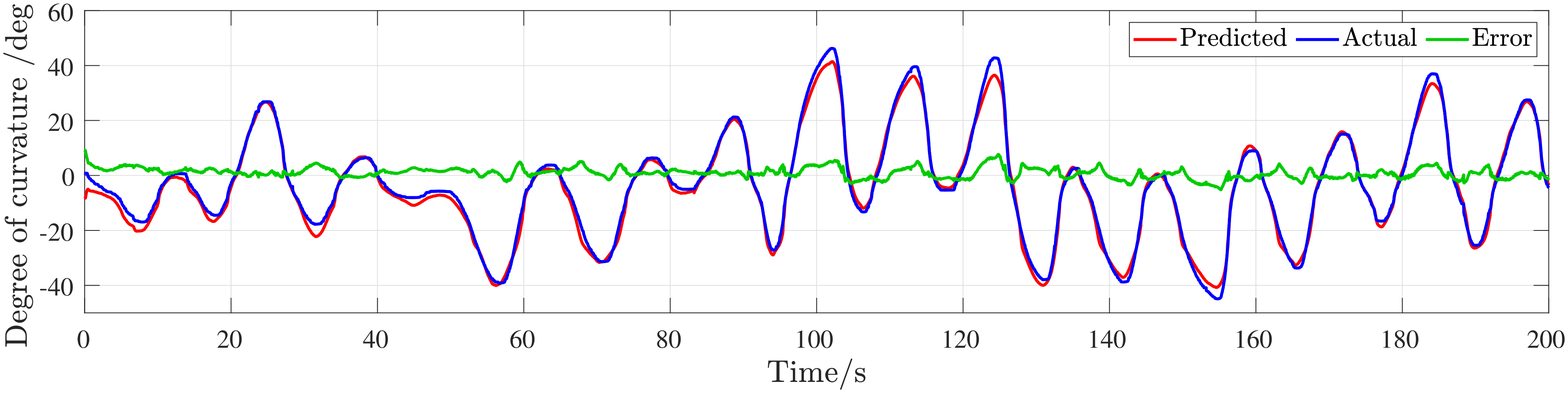}\\
		\vspace{5pt}
		\footnotesize{(c) Test set prediction from learned model}
		\vspace{-5pt}
	\end{tabular}
	\caption{Learned model test set performance: Bi-directional bending}
	\label{Fig:ML_model_both} \vspace{-12pt}
\end{figure}

For this case both \textit{compartments A} and \textit{B} were actuated, and the strain signals from both the sensing skins were used. The training data set was collected at a rate of $60$ Hz by conducting seven experiments using randomly generated actuation patterns at a varying rate of $1-4$ Hz. The cumulative length of the experiments was $30$ mins, resulting in a total data set that consisted of $107,090$ data points. This data set was then divided into a training set of $74,962$ points and two validation sets, $\alpha , \beta$-validation sets, of $16,064$ points each, from which the network was trained. The optimum number of hidden units for the LSTM layer was 30.

The degree of curvature estimation performance of the learned model was evaluated in real time for a random actuation pattern. The predicted degree of curvature by the LSTM network superimposed with the actual values for this experiment are illustrated in Fig.\ref{Fig:ML_model_both} along with the raw strain signals and the actuation signals. The RMSE for this test experiment was $1.95^{\circ}$.

\section{Experimental results} \label{sec:experiments}

In this section, we illustrate the efficacy of the integrated sensing strategy for dynamic tracking control of soft robots using the adaptive control framework (\ref{eq:adaptiveController})-(\ref{eq:adaptationgain}) for degree of curvature tracking. We report the results for both uni-directional and bi-directional bending. The tracking errors are computed and related to the degree of curvature ground truth measured by the MoCap system. 

Uncertainty was assumed in the segment mass, torsional stiffness and torsional damping. Thus the parameter vector was chosen as ${\Theta}_s = \left[\text{m}_{1}\text{L}_1^2, \: \text{K}_s, \: \text{D}_s\right]^\top$. The initial parameter estimates $\hat{\Theta}_s(0) = \left[0.6, \: 0.1, \:0.1\right]^\top$ were set different from the measured nominal values. The control gains were constant throughout the experiments and were set to $\Gamma = 1.2$, $\lambda = 3.2$, $K_{D} = 0.8$. 

\subsubsection{Uni-directional bending}
\begin{figure}[t]
	\centering
		\includegraphics[width=0.48\textwidth]{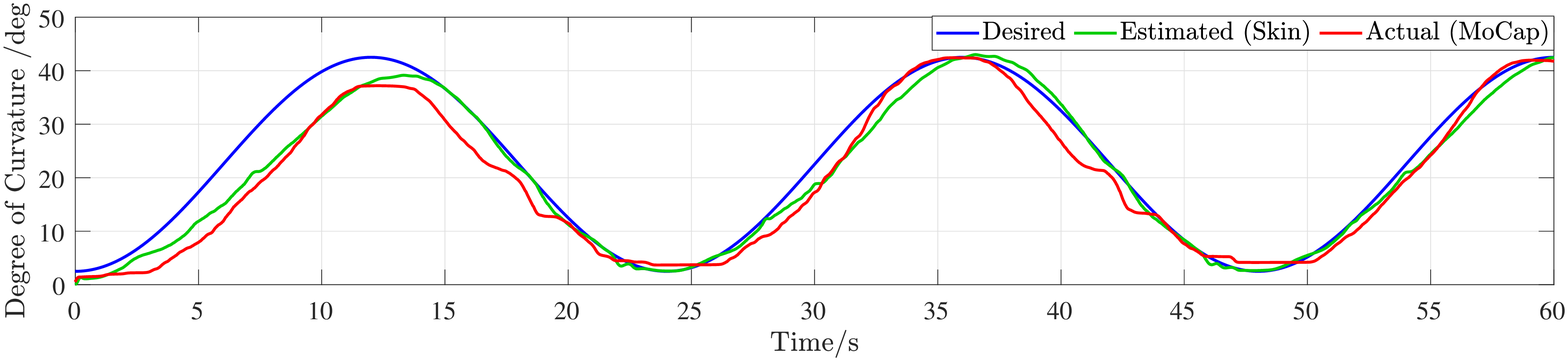}
		\vspace{-15pt}\\
	    \footnotesize{(a) Tracking performance}
	    \vspace{8pt}\\
		\includegraphics[width=0.48\textwidth]{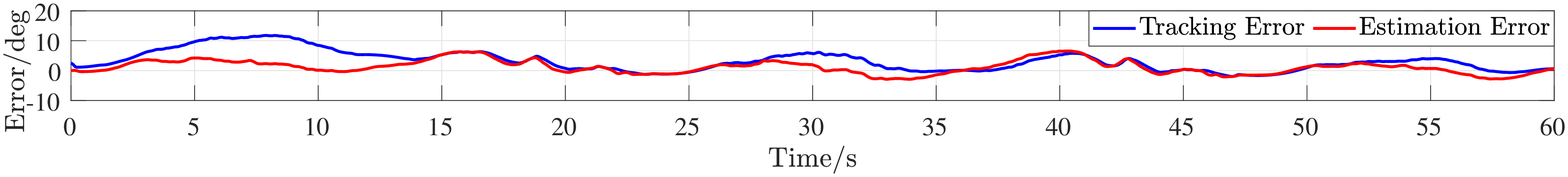} \vspace{-2pt}\\
		\footnotesize{(b) Error plot} \vspace{-5pt}
	\caption{Uni-directional low frequency target trajectory tracking}
	\label{Fig:single_low}
\end{figure}

\begin{figure}[t]
	\centering
		\includegraphics[width=0.48\textwidth]{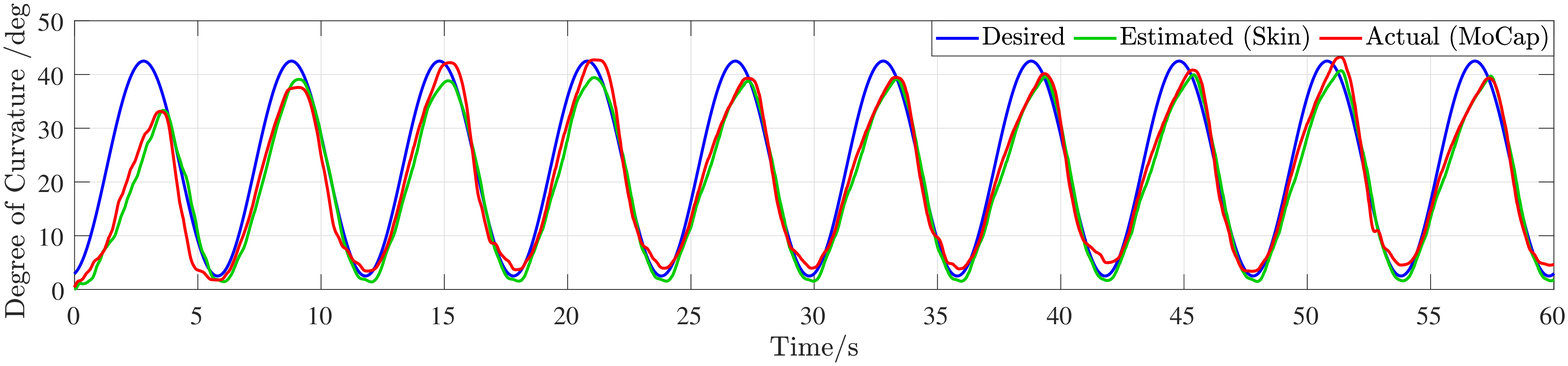}
		\vspace{-15pt}\\
	    \footnotesize{(a) Tracking performance} 		\vspace{8pt}\\
		\includegraphics[width=0.48\textwidth]{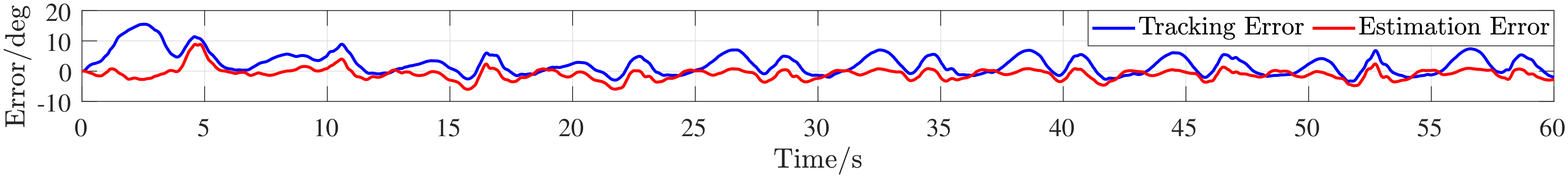} \vspace{-5pt}\\
		\footnotesize{(b) Error plot}  \vspace{-5pt}
	\caption{Uni-directional high frequency target trajectory tracking}
	\label{Fig:single_high_tracking} \vspace{-5pt}
\end{figure}

Here only the sensor strain signals from the skin retrofitted onto \textit{compartment A} and the actuator signals for \textit{compartment A} were used for degree of curvature estimation via the learned model, although both the compartments were allowed to be actuated. Two experiments were conducted, one with a low frequency target trajectory, and the other with a high frequency target trajectory. For the low frequency target, the desired degree of curvature was set to $q_d(t) = (\pi/8) - (\pi/9)\cos{\left(\pi t/12 \right)}$.
The results are shown in Fig.\ref{Fig:single_low} wherein the tracking RMSE was $4.35^{\circ}$, and the estimation RMSE was $2.78^{\circ}$. The results for the high frequency trajectory tracking are shown in Fig.\ref{Fig:single_high_tracking}, where $q_d(t) = (\pi/8) - (\pi/9)\cos{\left(\pi t/3 \right)}$.
In this case the tracking and estimation RMSE was found to be $4.09^{\circ}$ and $2.27^{\circ}$ respectively.

\subsubsection{Bi-directional bending}
\begin{figure}[t]
	\centering
		\includegraphics[width=0.48\textwidth]{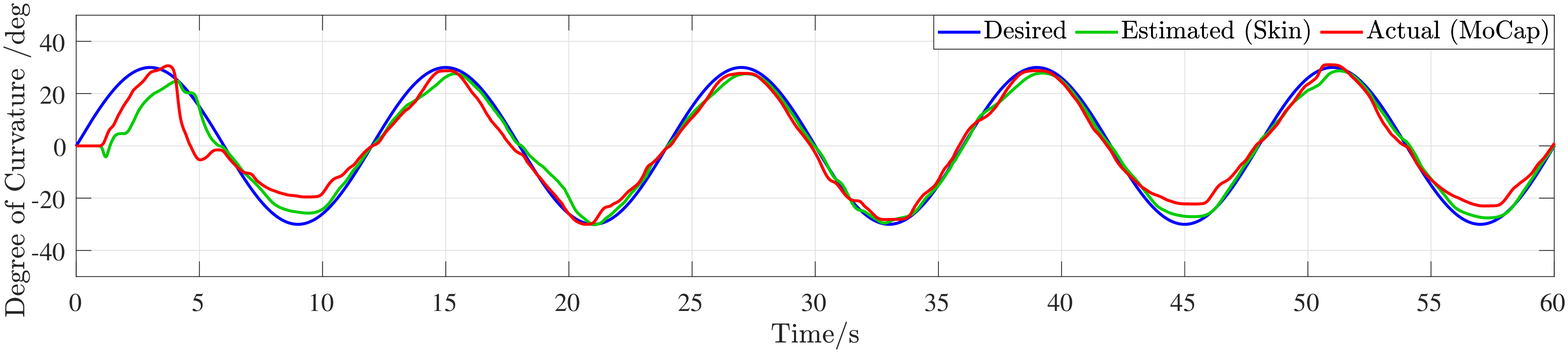} \vspace{-15pt}\\
	    \footnotesize{(a) Tracking performance}
	    \vspace{8pt}\\
		\includegraphics[width=0.48\textwidth]{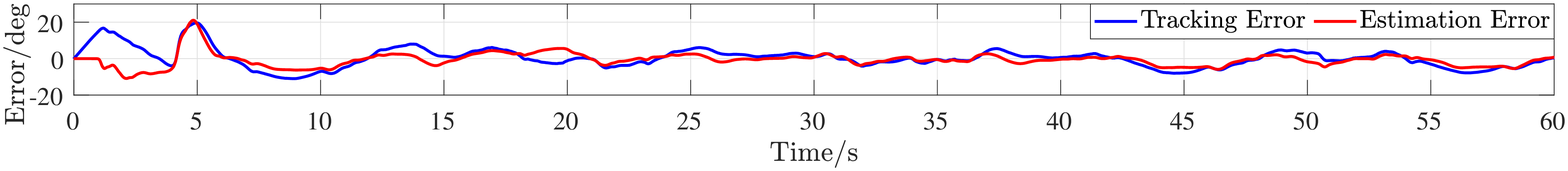}  \vspace{-2pt}\\
		\footnotesize{(b) Error plot}  \vspace{-5pt}
	\caption{Bi-directional low frequency target trajectory tracking}
	\label{Fig:both_low}
\end{figure}

\begin{figure}[t]
	\centering
		\includegraphics[width=0.48\textwidth]{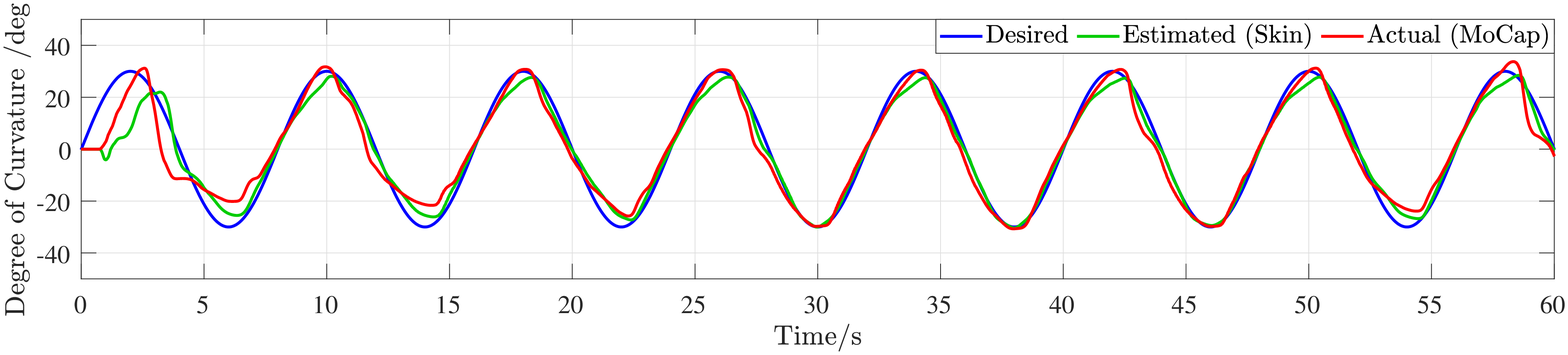}
		\vspace{-15pt}\\
	    \footnotesize{(a) Tracking performance} 		\vspace{8pt}\\
		\includegraphics[width=0.48\textwidth]{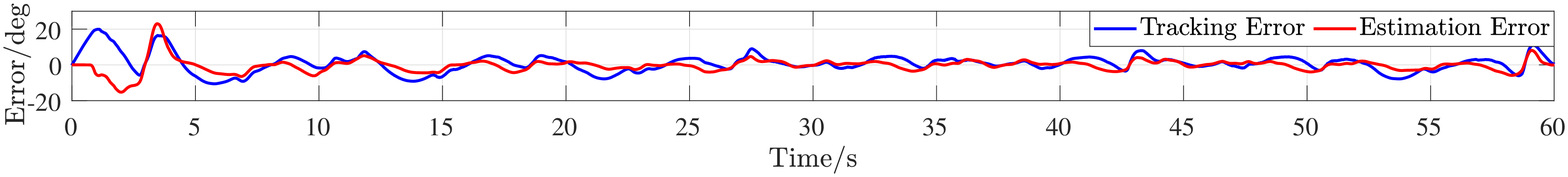}  \vspace{-2pt}\\
		\footnotesize{(b) Error plot}  \vspace{-2pt}
	\caption{Bi-directional high frequency target trajectory tracking}
	\label{Fig:both_high} \vspace{-5pt}
\end{figure}

Here both \textit{compartments} were actuated, and strain signals from both the sensing skins were used for degree of curvature estimation via the learned model for bi-directional bending. Two experiments were conducted, one with a low frequency target trajectory and one with a high frequency target trajectory. For the low frequency target, the desired degree of curvature was set to $q_d(t) = (\pi/6)\sin{\left(\pi t/6 \right)}$.
The results are shown in Fig.\ref{Fig:both_low}, in which the tracking and estimation RMSE was found to be $5.05^{\circ}$ and $3.79^{\circ}$ respectively. The results for the high frequency trajectory tracking are shown in Fig.\ref{Fig:both_high}, where $q_d(t) = (\pi/6)\sin{\left(\pi t/4 \right)}$. 
 Here the tracking RMSE was $5.10^{\circ}$ and the estimation RMSE was $3.73^{\circ}$.

\section{Discussion} \label{sec:discussion}
The experimental results exhibit the successful utilization of the retrofitted soft sensing skin for the degree of curvature estimation for adaptive tracking control of a desired curvature trajectory. The uni-directional bending illustrates the use of a single soft sensing skin for degree of curvature estimation when the soft robot only bends in a certain direction. This capability is useful for sensing and estimation of soft segments, such as in wearable robots, that have only a single compartment and only bend in a one direction. The bi-directional experiments demonstrate the use of two sensing skins retrofitted on the two compartments of the soft robot for curvature estimation. The bi-directional bending is especially important in soft robot manipulation.

Considering both the uni-directional and bi-directional bending, the capability of the integrated sensing skins to estimate the curvature for slow as well as fast manipulations are shown, and the fast response of the sensors are reflected in the satisfactory tracking of the target trajectory. In the starting of the experiments the higher tracking error maybe due to uncertain parameters which in time gets better due to parameter adaptation.

\section{Conclusion and future work} \label{sec:conclusion}
In this paper, we demonstrated the use of integrated sensing for dynamic control of soft robots under the piecewise constant curvature modeling hypothesis. The soft sensing skins proposed in this work could be retrofitted to many soft robots, and the degree of curvature estimation can be learned using an LSTM network, only requiring the strain signals from the sensing skin and the actuator inputs. Moreover, an adaptive controller was designed to track a desired degree of curvature trajectory. The satisfactory degree of curvature tracking using the adaptive controller for low and high frequency target trajectories demonstrates that the proposed soft skins are capable of estimating the degree of curvature robustly for inclusion in a dynamic control framework. 

The current work was only focused on free bending of a planar piecewise constant curvature soft robot. Subsequent work will be on developing and using the stretchable sensors for 3D soft robots under external wrenches while relaxing the piecewise constant curvature assumption.

\section{Acknowledgements}
The authors would like to thank the Robotics Realization Lab of the Maryland Robotic Center and the lab manager Dr. Ivan Pensky for helping to conduct physical experiments.
\\

\bibliographystyle{IEEEtran}
\bibliography{ICRA_soft_robot}

\end{document}